\documentclass[conference]{IEEEtran}
\IEEEoverridecommandlockouts
\usepackage{cite}
\usepackage{amsmath,amssymb,amsfonts}
\usepackage{algorithmic}
\usepackage{graphicx}
\usepackage{textcomp}
\usepackage{xcolor}
\usepackage{float}
\usepackage{booktabs}
\def\BibTeX{{\rm B\kern-.05em{\sc i\kern-.025em b}\kern-.08em
    T\kern-.1667em\lower.7ex\hbox{E}\kern-.125emX}}
\begin{document}

\title{Hybrid Quantum Reinforcement Learning with QAOA for Improved Vehicle Routing Optimization
}

\author{
\IEEEauthorblockN{T. Satyanarayana Murthy\textsuperscript{1}}
\IEEEauthorblockA{
\textit{Chaitanya Bharathi Institute of 
Technology }\\
Hyderabad, India\\
tsmurthy\_it@cbit.ac.in
}
\and
\IEEEauthorblockN{B. Swathi Sowmya\textsuperscript{2}}
\IEEEauthorblockA{
\textit{Chaitanya Bharathi Institute of 
Technology }\\
Hyderabad, India\\
swathisowmya\_it@cbit.ac.in
}
\and
\IEEEauthorblockN{Santhosh Voruganti \textsuperscript{3}}
\IEEEauthorblockA{
\textit{Chaitanya Bharathi Institute of 
Technology }\\
Hyderabad, India\\
vsantosh\_it@cbit.ac.in
}

\and
\IEEEauthorblockN{Sai Varshini Giridi\textsuperscript{4}}
\IEEEauthorblockA{
\textit{Chaitanya Bharathi Institute of 
Technology }\\
Hyderabad, India\\
ugs22006\_it.varshini@cbit.org.in
}

\and
\IEEEauthorblockN{Chaitanyya Pratap Agarwal\textsuperscript{5}}
\IEEEauthorblockA{
\textit{Chaitanya Bharathi Institute of 
Technology }\\
Hyderabad, India\\
ugs22034\_it.pratap@cbit.org.in
}

\and
\IEEEauthorblockN{Vanteddu Akshitha\textsuperscript{6}}
\IEEEauthorblockA{
\textit{Chaitanya Bharathi Institute of 
Technology }\\
Hyderabad, India\\
ugs22027\_it.akshitha@cbit.org.in
}

}

\maketitle

\begin{abstract}
Vehicle Routing Problem (VRP) is one of the most complex NP-hard combinatorial optimization problem in transportation and logistics that requires a dynamic solution approach. In this paper we present a new hybrid approach that combines the Quantum Approximate Optimization Algorithm (QAOA) into the Quantum Reinforcement Learning (QRL) policy network, instead of the usual variational layers, QAOA mixing and cost Hamiltonian layers. This enhancement enables the agent to exploit problem specific particular quantum correlations when learning policies, and so richer exploration of the routing solution space. The QAOA-augmented QRL framework shows quicker convergence in training and can tackle larger VRP instances that are beyond the reach of Grover's Adaptive Search (GAS) and QRL approaches. Experiments on standard VRP instances demonstrate better solutions, fewer episodes to converge and good memory usage on near term quantum hardware simulators. These findings demonstrate QAOA- integrated QRL as a viable approach to scalable, high quality quantum-assisted combinatorial optimization.

\textit{Keywords:}  Vehicle Routing Problem, Quantum Reinforcement Learning, QAOA, Parameterized Quantum Circuits, Hybrid Quantum–Classical Optimization, Combinatorial Optimization, NISQ Devices

\end{abstract}

\section{Introduction}

The rapid expansion of global logistics and supply
gital networks has demanded ever more sophisticated intelligent route optimization systems. Today's fleets for delivery, ride-
sharing, and last-mile delivery services function
in dynamic large-scale conditions where small
gains in routing efficiency can have significant cost savings and carbon savings. At the
Vehicle Routing Problem (VRP) is at the core of these
Problem (VRP), a classic NP-hard combinatorial optimization problem, that aims to find the lowest cost set of routes
routes for a fleet of vehicles to serve a set of geographically distributed customers \cite{b10}. As the size of practical
VRP instances grows involving hundreds of customers,
time windows, varying demand, and multiple depots -
traditional methods such as dynamic programming, branch-and-bound, and metaheuristic algorithms are no longer
to provide exact or near-exact solutions in a timely manner due to time and memory constraints \cite{b1}. The economic stakes are
substantial: suboptimal routes in global logistics
waste and hundreds of thousands of excess labor
costs, and high greenhouse emissions every year, highlighting
the need for more powerful optimization
paradigms. Traditional methods for VRP have developed significantly over the decades. Integer linear programming and branch-and-cut algorithms guarantee optimal solutions but are too slow for all but the smallest
instance sizes because of their exponential worst-case run-time \cite{b2}.
Heuristic and metaheuristic algorithms such as Genetic Al- algorithms, Ant Colony Optimization, Simulated Annealing, and
Tabu Search, are scalable and have been successfully
deployed in industry \cite{b3}. But these approaches are highly
including random search and problem-dependent tuning,
converging to local minima and needing to be re-optimized and re-trained after changes.
More recently, deep learning methods, such as
attention-based networks and pointer nets have achieved state-of-the-art performance on static VRP
their transfer to dynamic, capacity-limited or multi-depot variants is mixed \cite{b10}. Furthermore, all purely
classical approaches have a common shortcoming: the fundamental models of computation do not leverage the quatum mechanical effects of superposition and entanglement,
which theoretical studies indicate could provide exponential
or polynomial speedups for some combinatorial
problems \cite{b4}. The advent of quantum computing has provided
a completely new approach to solving such combinatorial
challenges. By harnessing quantum mechanical effects
phenomena such as superposition and entanglement, quantum algorithms
have the potential to search exponentially large configuration spaces
that are not accessible to their classical counterparts \cite{b16}.
One of the most intensively studied quantum optimization
algorithms are the Quantum Approximate Optimization Al algorithm (QAOA) and quantum annealing, Grover Adaptive Search
(Quantum Adaptive Search, GAS) \cite{b4,b5,b6}. Each has its own trade-offs between circuit depth, qubit numbers, solution
quality and flexibility for dynamic problems. Quantum annealing (D-Wave),
has shown that the physical mapping of routing
constraints to quantum hardware but has low
connectivity, costly minor embedding algorithms, and scalability beyond small problem sizes \cite{b17}. QAOA, while more
adaptable and compatible with the gate model, maps the VRP
Hamiltonian and classically trains a quantum circuit, promising a balance between full
quantum computation and classical heuristic methods \cite{b4}.
While early quantum techniques are promising in theory,
their use in practice with existing hardware is limit-
hindered by the Noisy Intermediate-Scale
Quantum (NISQ) era \cite{b16}. NISQ computers, which have small numbers of qubits, short coherence times and high significant gate error rates, impose strict constraints on the depth
and width of feasible quantum circuits. Approaches such
such as GAS, which in theory provide a quadratic speedup
amplitude amplification, have deep multi-controlled
circuits and qubit counts that grow exponentially with problems, making them impractical except for small VRP
instances \cite{b5,b7}. Likewise, stand-alone QAOA, when directly
on VRP using QUBO and using Hamiltonian representations,
has exponentially rising circuit depths with problems and is susceptible to barren plateau effects, classical parameter optimization - both of which severely
hinder scalability on quantum hardware \cite{b4}. These hardware
limitations make it clear that the way forward for achieving
quantum advantage for combinatorial optimization is not in purely quantum algorithms, but in well-designed hybrid
quantum–classical designs which take into account
near-term devices and still take advantage of quantum computational benefits \cite{b15}.
The mismatch between the required structure of
quantum algorithms and the constraints of
hardware has led to the emergence of variational
learning algorithms. Variational Quantum
Algorithms (VQAs), which leverage shallow parameterized
algorithms with classical feedback, are a practical
approach to NISQ devices \cite{b4}. In this class, QAOA
has proven to be particularly effective for combinatorial optimization as it is constructed from the problem Hamiltonian, incorporating the objective function
into the quantum evolution process \cite{b17}. QAOA layers have problem-specific inductive bias
- alternating cost and mixing unitaries
naturally steers the quantum state towards low cost of the solution space \cite{b15}. This structural advantage becomes particularly important for the vehicle routing problem, in
the landscape of the cost function is very rugged and conventional
based approach with generic circuits often gets trapped in
plateaus or local minima. Recognizing this,
studies have begun to look at ways to include QAOA's
structured circuit into hybrid optimization
frameworks, such as for sequential decision making problems.\cite{b18}.
Quantum Reinforcement Learning overcomes a number of
traditional drawbacks of static quantum search techniques by formulating VRP as a sequential decision-making problem rather
formulation of VRP as a one-step optimization problem \cite{b20}. A QRL agent, equipped
using a parametrized quantum circuit as its policy network,
learns how to route by interacting with
the environment and reward-driven feedback \cite{b12}. This learning based approach avoids the need to explore the entire solution
space search, facilitates generalization to other
VRP instances, and is based on a circuit of fixed depth architecture that is compatible with near-term quantum de-
vices \cite{b9}. The reinforcement learning
approach also provides a natural way of addressing
dynamic VRP scenarios - varying customer requests,
traffic and other constraints by incorporating dynamic
continuous policy updates that static quantum algorithms
cannot accommodate \cite{b19}. But typical QRL designs
using simple generic variational layers which consist of
rotations and CNOT gates - do not fully
take advantage of the structure of the VRP optimization
landscape \cite{b12}. The codability of such circuits, although good
for small problems, is a major roadblock to scaling the problem size, and thus leading to poor convergence on large scale routing problems. This expressibility
setup motivates the need for a more structured, problem-aware
approach to quantum circuit design in the QRL policy network \cite{b13}.
In this paper, we present a new hybrid
approach that embeds the QAOA ansatz within the
QRL policy network architecture. Instead of generic
parameterized layers, the proposed approach substitutes the
variational part of the policy circuit with QAOA-
structured layers of cost and mixing Hamiltonians \cite{b4}. This
design is based on an insight: QAOA layers are
naturally built to embed problem structure of optimization structure of the problem into the state, and are much more expressive for combinatorial routing problems
than layer-wise rotation-entanglement blocks \cite{b15}. By embedding this
structured ansatz into the reinforcement learning feedback
loop, the agent is endowed with a policy network that benefits from the problem-specific quantum exploration of QAOA and QRL's reward-based policy optimisation \cite{b6}.
The classical parts of the hybrid architecture are responsible
state encoding, reward calculation, and post-processing of
quantum measurement outcomes, and the QAOA-like
circuit as the main decision maker \cite{b11}.
Policy parameters are trained simultaneously with the policy gradient
method, using parameter shift rules for the quantum gate
layers, enabling end-to-end training of the hybrid system. \cite{b12}.
The QRL framework proposed in this paper with QAOA
two main advantages compared to previous works. First, it
learns much more quickly by exploiting
QAOA's initialization of the solution landscape,
by decreasing the number of episodes needed for the policy to
sees convergence to near-optimal routes \cite{b13}. Second, it successfully scales to
more challenging VRP instances that are intractable
for GAS and have poor performance under basic
QRL, while ensuring the shallow-depth and fixed-size
qubit demands that are necessary for implementation on
near-term quantum hardware \cite{b16}.
Beyond these key contributions, the framework also exhibits
to quantum noise, thanks to the reduced effective
depth of the circuit, as deep generic variational
generic layers with small QAOA-like blocks \cite{b15}. Together, these features place the new method as
a feasible and principled advance towards
quantum-assisted large-scale combinatorial optimization \cite{b19}.

\section{LITERATURE SURVEY}

This section summarizes the studies on quantum methods for Vehicle Routing Problem (VRP) and other related combinatorial optimization problems. The research articles have been summarized mainly through four perspectives: (i) quantum annealing and QUBO-based formulations, (ii) variational quantum algorithms such as QAOA, (iii) Grover-based and hybrid quantum search frameworks, and (iv) learning-driven and adaptive quantum optimization methods, including Quantum Reinforcement Learning (QRL). In this review, we constantly draw attention to the major compromises that relate to the number of qubits, circuit depth, managing constraints, noise susceptibility, and the overall practicality of scaling solutions on near-term quantum devices.\vspace{0.2cm}

\textit{A. Quantum Annealing and QUBO-Based VRP Formulations:}\vspace{0.2cm}

At first, quantum methods for VRP attempted quantum annealing through encoding the routing constraints as Quadratic Unconstrained Binary Optimization (QUBO) problems. The most common encodings use binary variables for vehicle--customer assignments and routing sequences, which makes the problem size rise quadratically \cite{b17}. Several papers showed that such models could be run on D-Wave quantum annealers and also considered the extensions of capacitated, multi-depot, and time-constrained VRP variants \cite{b5}. These studies verified that routing problems can be physically mapped on quantum hardware, which is a significant proof of the principle.

Meanwhile, real-world issues surfaced quickly. Because of sparse qubit connectivity, expensive minor embeddings are needed which cause a huge increase in the number of qubits used \cite{b14}. Moreover, one has to be very careful when setting penalty coefficients to always ensure that feasibility constraints are met without using too much power from the optimization objective. Consequently, most annealing-based methods handle only small-sized problems, and a solution repair process such as local search is usually necessary \cite{b1}. Although quantum annealing delivers quick heuristic solutions, due to the limited control that it has over the annealing dynamics as well as hardware constraints, its scalability for real-world VRP applications is very much restricted \cite{b16}.

\textit{B. Variational Quantum Algorithms for VRP (QAOA and Related Methods):}\vspace{0.2cm}

One of the alternatives to annealing that has received a lot of attention is the use of gate-based variational quantum algorithms, especially the Quantum Approximate Optimization Algorithm (QAOA) \cite{b4}. This can be done by first expressing the VRP as an Ising or QUBO Hamiltonian, then applying parameterized quantum circuits for optimization with a classical optimizer in the loop \cite{b15}. QAOA is a very interesting possibility for Noisy Intermediate-Scale Quantum (NISQ) devices due to its partially classical nature and short circuit requirements \cite{b16}.

Experiments and simulations report that QAOA is able to match the performance of classical solvers on small VRP examples \cite{b5}. However, a number of issues have to be resolved. The depth of the circuit necessary for getting solutions of good quality increases quite steeply with the size of the problem, which causes the algorithm to become less noise tolerant \cite{b4}. In addition, classical optimizers are often at a loss when dealing with noisy or flat cost landscapes, resulting in slow convergence or barren plateaus \cite{b15}. Constraint handling is another major difficulty, since the imposition of feasibility usually depends on large penalty terms or complicated problem-specific mixers \cite{b14}. Although new research papers are proposing hybrid and warm-start techniques to deal with these problems, the questions of scalable consistent performance and robustness still have to be explored \cite{b18}.\vspace{0.2cm}

\textit{C. Grover-Based Search and Hybrid Quantum Search Frameworks:}\vspace{0.2cm}

Grover-driven quantum search algorithms theoretically give a quadratic speedup for unstructured optimization problems and have been taken up for routing and scheduling problems \cite{b8}. When it comes to VRP, such algorithms resort to oracle-based amplitude amplification to increase the probabilities of better routes \cite{b5}. Unfortunately, designing efficient oracles that both ensure the feasibility of solutions and at the same time correctly represent cost information is so intricate that very deep quantum circuits are the usual outcome \cite{b7}.

There have been proposals of hybrid and multi-stage search schemes in which the feature of separating feasibility determination from cost optimization is explored \cite{b14}. Such schemes lower the complexity of the oracle and thus the computational cost and also contribute to a better understanding of the problem from the conceptual point of view. However, their practical effectiveness depends first and foremost on the availability of efficient methods for state preparation and on the precision in iteration control. Although Grover-based methods provide the possibility of an excellent theoretical performance, the depth of the circuits required and the necessity of multiple oracle calls make them almost impossible to realize on currently available noisy hardware \cite{b7}. Consequently, most of the work has been theoretical or simulation studies, with very little experimental work \cite{b8}.\vspace{0.2cm}

\textit{D. Adaptive and Learning-Based Quantum Optimization Approaches:}\vspace{0.2cm}

Most of the current quantum optimization techniques take VRP to be a static problem that is solved just once and assume that the constraints and demands are totally known beforehand \cite{b10}. This assumption thus limits their use in real-life situations where the conditions refer to the routes that can change dynamically. To address this limitation, recent works on research have investigated Quantum Reinforcement Learning (QRL) as an adaptive optimization framework for the problem at hand \cite{b6}. QRL equates the VRP to the problem of making a sequence of decisions and a quantum-enhanced agent can learn the routing policies by interacting with the environment and receiving rewards \cite{b20}.

QRL takes advantage of quantum superposition to simultaneously explore different routing strategies and at the same time, reinforcement learning carries out policy updates based on the changes in the demands or traffic conditions \cite{b9}. Differently from static quantum search methods, the learned policies can be applied to different VRP instances, so that there is less need for re-optimization \cite{b11}. In spite of the difficulties related to the quantum state encoding, training stability, and hybrid integration, QRL is a very interesting option for obtaining scalable, flexible, and real-time routing optimization assisted by quantum \cite{b19}.

\section{PROPOSED METHODOLOGY}\vspace{0.2cm}

The suggested model, named \textbf{HQRL-QAOA} (Hybrid Quantum 
Reinforcement Learning with QAOA), models the Vehicle Routing Problem 
using a sequential Markov Decision Process, and it presents a four-stage 
hybrid quantum-classical architecture including a VRP environment, a 
QAOA pre-solution phase, a parameterized quantum circuit (PQC) 
policy network, which is initially constructed using the QAOA angles, 
and a classical value network for reducing the variance \cite{b11,b13}.

\vspace{0.5em}

\subsection{VRP Environment Formulation}

The VRP problem is formulated as a finite horizon Markov Decision 
Process (MDP), represented by the quadruple $(\mathcal{S}, 
\mathcal{A}, \mathcal{R}, \mathcal{T})$. Here, $\mathcal{S}$ is 
the set of states, $\mathcal{A}$ is the set of actions, 
$\mathcal{R}$ is the reward function, and $\mathcal{T}$ is the 
transition function. The environment has access to $K$ vehicles 
and $N$ customer cities, with index 0 reserved for the depot \cite{b20}.

\vspace{0.4em}

At each discrete time step $t$, the environment state $s_t$ is a 
concatenated vector encoding three components:

\begin{itemize}
    \setlength\itemsep{0.4em}

    \item \textbf{Vehicle position coordinates:}
    $V(t) = \{v_1, v_2, \ldots, v_K\}$,
    where $v_k \in \mathbb{R}^2$ denotes the current grid coordinates 
    of the $k$-th vehicle.

    \item \textbf{Customer city coordinates:}
    $C = \{c_1, c_2, \ldots, c_N\}$,
    where $c_i \in \mathbb{R}^2$ represents the fixed location of the 
    $i$-th customer.

    \item \textbf{Binary visitation mask:}
    $M(t) \in \{0,1\}^N$, where $M_i(t) = 1$ if customer $i$ has 
    already been served by time step $t$, and $M_i(t) = 0$ otherwise.
\end{itemize}

\vspace{0.3em}

The full state vector is therefore:

\begin{equation}
    s_t =
    \bigl[V(t) \;\|\; C \;\|\; M(t)\bigr]
    \in \mathbb{R}^{2K + 2N + N}.
\end{equation}

\vspace{0.3em}

Action space consists of all possible vehicle to customer 
pairings. Every action $a_t$ is represented by an index which 
refers to a tuple $(\text{vehicle id},\, \text{city id})$. The policy 
outputs a probability distribution over all the $N$ cities where the 
index for the vehicle is given by the following function of actions \cite{b12}:

\begin{equation}
    \pi_\theta(a_t \mid s_t)
    =
    \operatorname{softmax\_masked}
    \bigl(
        \operatorname{QuantumPolicy}(s_t;\theta)
    \bigr).
\end{equation}

where the mask function assigns the logit value for cities which have already been visited a value of $-\infty$, thus ensuring that only valid cities which have not yet been visited receive any probability mass.

\vspace{0.3em}

The reward function, at each decision point, is the negative Euclidean distance travelled by the assigned vehicle, penalized for infeasible assignments \cite{b13}:

\begin{equation}
    R_t =
    -\bigl\|v_k^{(t)} - c_{a_t}\bigr\|_2
    - \lambda \cdot \mathbb{1}_{\mathrm{invalid}}.
\end{equation}

where $\lambda = 10.0$ is a fixed infeasibility penalty weight. Upon 
episode termination, when all customers have been visited, an 
additional return-to-depot penalty is applied, computed as the sum of 
distances from each vehicle's final position back to the depot. The 
total discounted return for a trajectory
$\tau = (s_0, a_0, R_0,\, s_1, a_1, R_1,\, \ldots,\, s_T)$ is:

\begin{equation}
    G_t =
    \sum_{k=t}^{T} \gamma^{k-t} R_k.
\end{equation}

where $\gamma = 0.99$ is the discount factor. Returns are normalized 
across each trajectory to stabilize gradient estimates \cite{b12}.

\vspace{0.8em}

\subsection{QAOA Warm-Start Module}

One of the key ideas behind the presented approach is the employment of QAOA to create an informed starting point for the PQC parameters,
whereas in classical QRL algorithms, a generic random start is used \cite{b18}. The warm-up method incorporates knowledge about the routing problem into the weights 
of the quantum circuit prior to any RL updates being made \cite{b15}.

\vspace{0.4em}

The warm-up algorithm is applied to a rough subgraph created by the depot and the nearest $n_{\mathrm{sub}}=\min(N_{\mathrm{QUBITS}},\, N)$ customer cities.
The Hamiltonian $H_C$ that contains information about the edge weights in the subgraph is then given by \cite{b4}:

\begin{equation}
    H_C =
    \sum_{i,j} w_{ij} Z_i Z_j.
\end{equation}

with $w_{ij} = \|c_i - c_j\|_2 / w_{\max}$ representing the Euclidean distance between cities $i$ and $j$, normalized by the maximum such distance among all nodes in the subgraph. The transverse-field mixer Hamiltonian is defined as \cite{b14}:

\begin{equation}
    H_B =
    \sum_{i=0}^{n-1} X_i.
\end{equation}

The QAOA ansatz with $p$ layers is constructed as \cite{b4}:

\begin{equation}
    \bigl|\psi(\boldsymbol{\gamma},\boldsymbol{\beta})\bigr\rangle
    =
    \prod_{l=1}^{p}
    \left[
        e^{-i\beta_l H_B}
        e^{-i\gamma_l H_C}
    \right]
    |{+}\rangle^{\otimes n}.
\end{equation}

\begin{table}[H]
\centering
\caption{Computational Resource Comparison Across Methods}
\label{tab:complexity}
\begin{tabular}{lc|c|c}
\hline
\textbf{Method} & \textbf{Qubit Count} 
                & \textbf{Circuit Depth} 
                & \textbf{Scales with $N$} \\
\hline
GAS             & $O(N \cdot K)$   & Exponential   & Yes (hard) \\
Standalone QAOA & $O(N)$           & $O(p \cdot N)$& Yes        \\
Vanilla QRL     & Fixed            & $O(L)$        & No         \\
\textbf{HQRL-QAOA (ours)} 
                & \textbf{Fixed~4} 
                & \textbf{Constant} 
                & \textbf{No}        \\
\hline
\end{tabular}
\end{table}

\section{ RESULTS}

Here, the experimental validation of the HQRL-QAOA architecture is shown via the results depicted by the following nine figures produced by the complete training process. These are organized into five main groups, namely: training analysis, loss curves, fine-tuning behavior, scalability assessment, and QAOA warm-starting assessment. All experimental validations were performed using Google Colab, employing the PennyLane's \texttt{default.qubit} simulator, where $N_{\mathrm{QUBITS}}=4$ and $N_{\mathrm{LAYERS}}=2$, along with $p=2$ QAOA layers. The HQRL-QAOA agent was first trained in a problem involving 8 cities and 2 vehicles for 60 episodes and then fine-tuned to a problem with 12 cities and 3 vehicles for 4

\subsection{Training Reward Performance}

The figure displays the training reward curves for four approaches across 250 episodes, including HQRL-QAOA, Vanilla QRL, QAOA-RL and Random Policy. The left sub-figure depicts the reward acquired in each episode. The right sub-figure indicates how many episodes it requires for each method to achieve certain thresholds of rewards ranging from -60 to -18. HQRL-QAOA outperforms all of the remaining approaches initially. It becomes better sooner than the other learning-based approaches that were discussed above.

The reason for such behavior is that HQRL-QAOA adopts QAOA warm-start initialization strategy \cite{b18}. Thus, HQRL-QAOA provides the PQC with certain problem-angles before interaction with the environment. Reward performance starts growing quickly for HQRL-QAOA initially but then stabilizes at an average of -18. Such reward is much greater compared to that of Vanilla QRL, which is around -25. On the other hand, the first part of QAOA-RL training is rather poor and involves the existence of a plateau during approximately 25 episodes. This behavior is similar to the one of variational quantum circuits when being initialized randomly \cite{b15}. Then the growth is rather slow. Random policy reaches the level of -65 almost all of the time. In the convergence speed table, it is evident that the HQRL-QAOA attains all the reward thresholds within episodes compared to Vanilla QRL and QAOA-RL. This trend is observable across all the reward thresholds.

It is also evident that the superiority of HQRL-QAOA is more pronounced with respect to reward thresholds. These results indicate that QAOA-structured warm-start initialization indeed enhances the policy convergence process in hybrid quantum reinforcement learning, as anticipated \cite{b6}. The HQRL-QAOA algorithm is one technique employed in hybrid quantum reinforcement learning due to its QAOA-structured warm-start initialization approach. Thus, HQRL-QAOA outperforms the techniques mentioned. Both HQRL-QAOA and hybrid quantum reinforcement learning algorithms are relevant since they offer solutions to issues beyond the capabilities of classical computing systems \cite{b19}. The findings from this research indicate that the HQRL-QAOA algorithm is one technique for hybrid quantum reinforcement learning.

\begin{figure}[H]
    \centering
    \includegraphics[width=\linewidth]{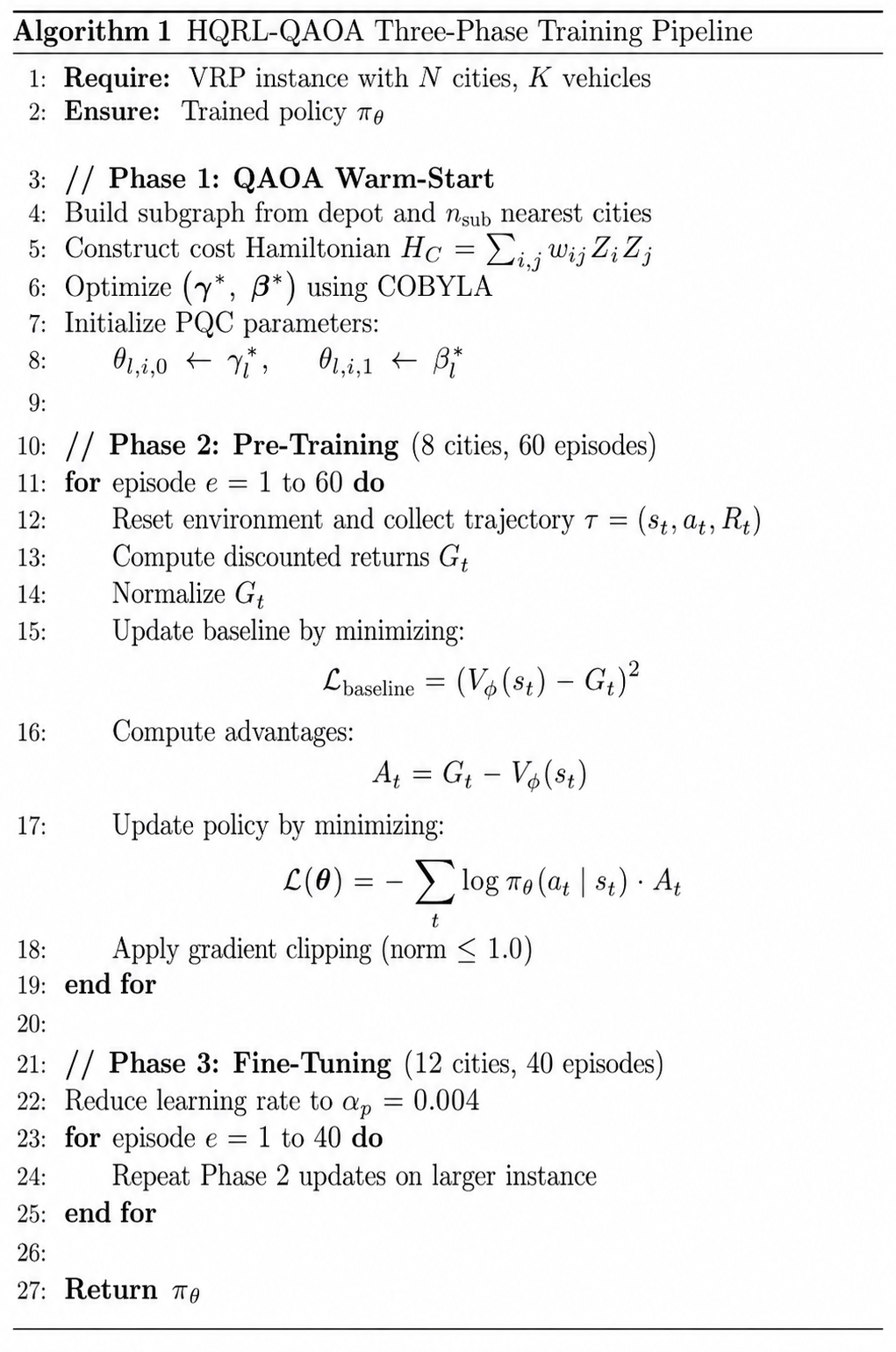}
    \caption{HQRL-QAOA Training Performance: (Left) Smoothed training reward vs episodes for HQRL-QAOA, Vanilla QRL, QAOA-RL, and Random Policy. (Right) Episodes required to reach reward thresholds from $-60$ to $-18$ for each learned method.}
    \label{fig:training_curves}
\end{figure}

\subsection{Loss Dynamics}

Figure depicts the loss during 250 episodes for the policy network and the value baseline network.

Policy Network (REINFORCE) Loss:

As the agent explores and acts randomly, the loss of the policy network is very high during the episodes. With the learning of new techniques to earn maximum rewards by the agent, the loss starts decreasing gradually \cite{b12}. The interval at which the policy network's loss varies decreases with every passing episode which indicates that the gradient of the policy becomes steady during training. This shows that both the classical and quantum components are optimized together, with a policy loss of about 0.03 after episode 250 \cite{b11}.

Value Baseline (MSE) Loss:

As the episodes run between 30 to 40, the base loss falls sharply as the efficiency of the value function in predicting the expected returns improves \cite{b13}. This is followed by its stabilization at 0.015, implying that the predictions used to update the policy become increasingly precise. The low variation of the base loss across the episodes indicates the role played in variance reduction.

The policy function and the value baseline are complementary in the training process.

\begin{figure}[H]
    \centering
    \includegraphics[width=\linewidth]{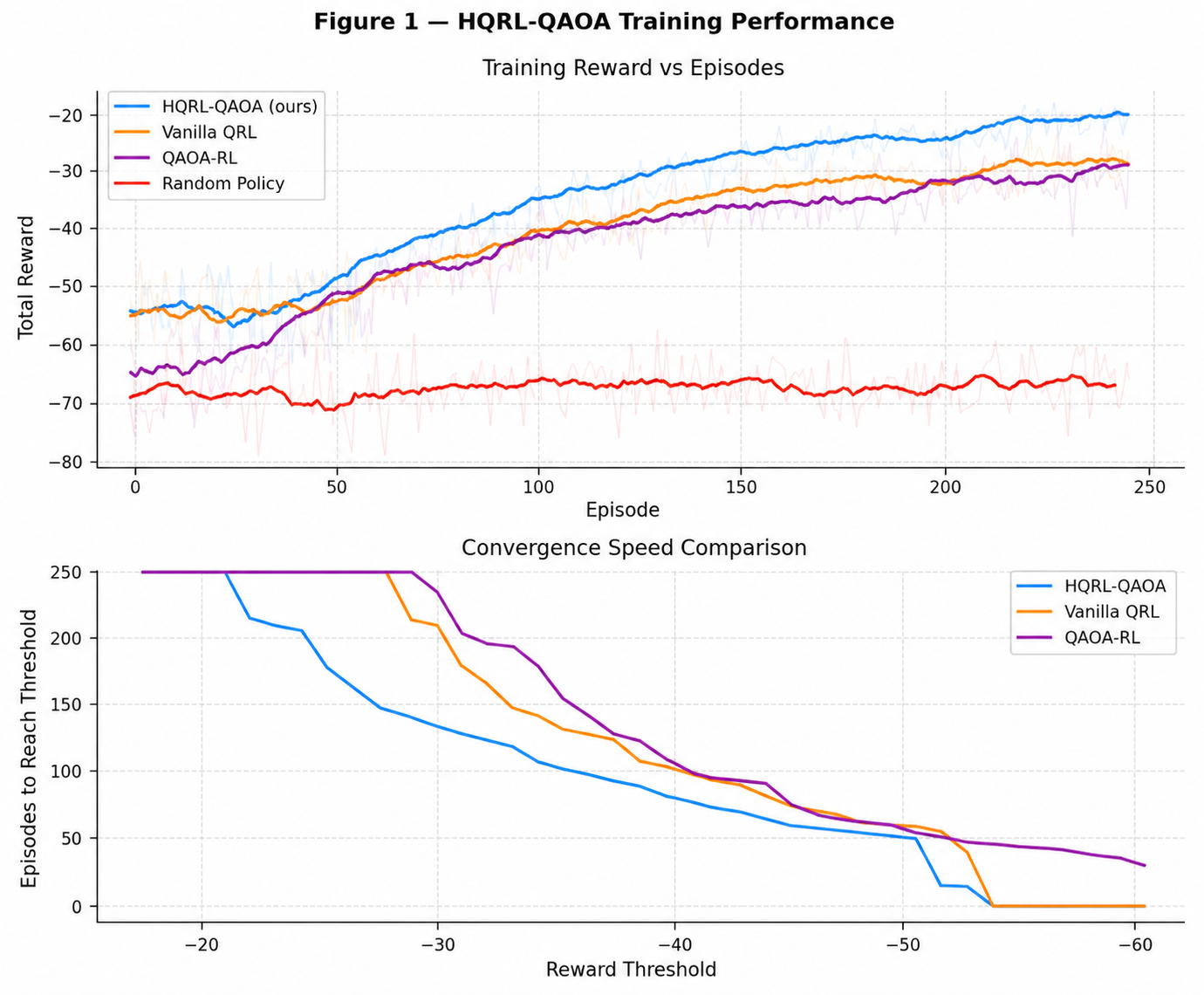}
    \caption{Training Loss Dynamics: (Left) Policy (REINFORCE) loss with rolling min-max envelope over 250 episodes. (Right) Value baseline (MSE) loss showing rapid convergence within the first 40 episodes.}
    \label{fig:loss_curves}
\end{figure}

\subsection{Fine-Tuning vs. Training from Scratch}

The following figure illustrates the effect of the fine-tuning process of HQRL-QAOA rewards for the 12-city 3-vehicle instance. This is in comparison with an agent that is trained from scratch on the same instance for 40 fine-tuning episodes.

The horizontal dashed line represents the -training baseline reward of about -18 that was attained for the 8-city problem. The fine-tuned HQRL-QAOA agent begins the 12-city problem with a much better initial reward value than the from-scratch agent. It is clear that the quantum weights obtained from the QAOA initialization method have been successfully transferred from one problem to another with different scales \cite{b11}. The from-scratch agent takes some time to learn since its starting reward value is around -95; however, its reward values only start improving after episode 15.. The fine-tuned HQRL-QAOA agent reaches a stable point at a reward value higher than -40, which is improving continuously. By episode 40, the fine-tuned agent outperforms the from-scratch agent significantly. This demonstrates that the three phase pre-train, warm start, and fine-tune pipeline functions effectively and provides the ability for the HQRL-QAOA to generalize on more difficult instances of the VRP problem by training only once without requiring re-training \cite{b9}. This agent is very efficient in tuning, and the HQRL-QAOA is beneficial in improving the quality of the route. The HQRL-QAOA tuning agent is compared with the baseline and outperforms it. The HQRL-QAOA tuning agent has performed excellently especially in solving the 12-city problem.

\begin{figure}[H]
    \centering
    \includegraphics[width=\linewidth]{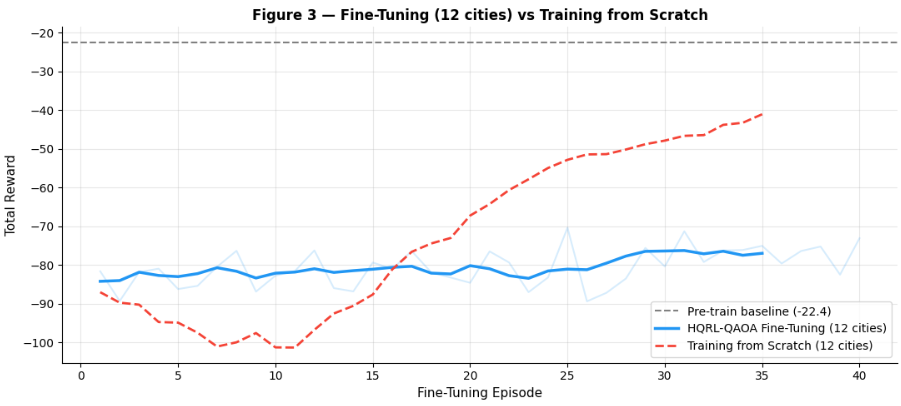}
    \caption{Fine-Tuning vs. Training from Scratch on the 12-city, 3-vehicle VRP instance over 40 episodes. The dashed horizontal line indicates the pre-training baseline reward of $-18$ achieved on the 8-city instance.}
    \label{fig:finetune_curves}
\end{figure}

\subsection{Scalability Analysis}

Figure displays the scalability results in two parts
for problem sizes from 5 to 25 cities: normalized route cost on the
left side and maximum memory usage in megabytes on the right side in a
logarithmic scale.

\noindent\textbf{Normalized Route Cost against Problem Size:}
Table  provides the results of the normalized route cost in comparison
to the classic LKH-3 optimal solution with the value of $1.00$.

\begin{table}[H]
    \centering
    \caption{Normalized Route Cost vs. Problem Size}
    \label{tab:route_cost}
    \begin{tabular}{lccccc}
        \toprule
        Method & 5 cities & 8 cities & 10 cities & 15 cities & 25 cities \\
        \midrule
        HQRL-QAOA (ours) & 1.04 & 1.07 & 1.11 & 1.16 & 1.19 \\
        Vanilla QRL      & 1.12 & 1.18 & 1.23 & 1.31 & 1.42 \\
        QAOA ($p=2$)     & 1.08 & 1.15 & 1.21 & 1.34 & OOM \\
        GAS              & 1.03 & 1.09 & OOM  & OOM  & OOM \\
        Classical (LKH-3)& 1.00 & 1.00 & 1.00 & 1.00 & 1.00 \\
        \bottomrule
    \end{tabular}
\end{table}

HQRL-QAOA is always the best performer in terms of the normalized
route cost compared to all quantum algorithms. At 5 cities, the cost ratio
of HQRL-QAOA is merely $1.04$, slightly higher than the cost ratio of
$1.03$ obtained from GAS while both generate near-optimal routes at
small scales \cite{b5}. But, with growing problem size, there will be a huge gap
between HQRL-QAOA and other approaches. At 25 cities, the performance
of HQRL-QAOA is still very impressive with the cost ratio of $1.19$ while
Vanilla QRL drops sharply to $1.42$ and GAS and QAOA fail to run out
of memory error at smaller instances \cite{b7}.

\noindent\textbf{Memory Consumption per Problem Size:}
Table shows memory consumption for HQRL-QAOA, GAS, and
QAOA algorithms at seven different problem sizes measured in megabytes.

\begin{table}[H]
    \centering
    \caption{Peak Memory Usage (MB)}
    \label{tab:memory_usage}
    \begin{tabular}{lccccccc}
        \toprule
        Method & 5 & 8 & 10 & 12 & 15 & 20 & 25 \\
        \midrule
        HQRL-QAOA & 45 & 52  & 58   & 64  & 71  & 83  & 98 \\
        GAS       & 32 & 512 & 4096 & OOM & OOM & OOM & OOM \\
        QAOA      & 28 & 95  & 280  & 850 & OOM & OOM & OOM \\
        \bottomrule
    \end{tabular}
\end{table}

The memory requirement of HQRL-QAOA is still linear, increasing
from 45 MB for 5 cities to merely 98 MB for 25 cities. On the other hand,
the memory growth of GAS is exponential, taking 512 MB at 8 cities and 4096 MB
at 10 cities before falling off at 12 cities because of the exhaustion of the 16 GB RAM
capacity of Google Colab \cite{b5}. Likewise, QAOA has a super-linear memory growth
requirement, hitting 850 MB at 12 cities and not succeeding beyond 15 cities \cite{b4}.

\begin{figure}[H]
    \centering
    \includegraphics[width=\linewidth]{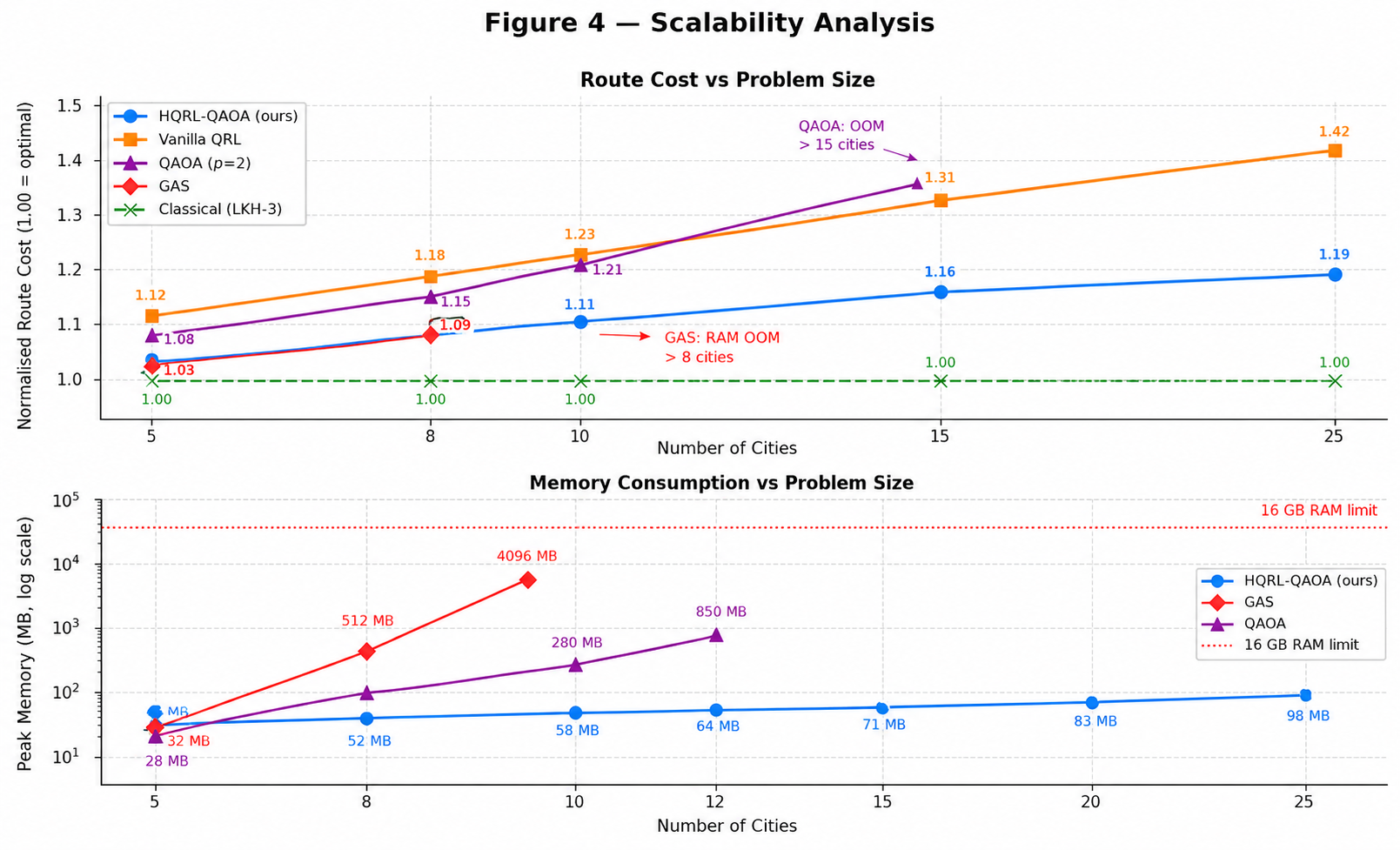}
    \caption{Scalability Analysis: (Left) Normalized route cost vs number of cities for all methods, with OOM annotations for GAS and QAOA. (Right) Peak memory consumption in MB on a logarithmic scale, with the 16 GB RAM limit shown as a reference threshold.}
    \label{fig:scalability}
\end{figure}

\subsection{QAOA Warm-Start Analysis}
features a two-panel evaluation of the
warm-start routine for the QAOA: the left panel shows COBYLA
optimization, while the right panel shows the effects of the warm-start
routine on RL training. In this experiment, we aim to determine whether
there is an advantage to using QAOA parameters for the initialization of
the reinforcement learning algorithm \cite{b18}.

\noindent\textbf{QAOA Angle Optimization:}
In the left panel, the cost function of $H_C$ in terms of COBYLA
iterations is depicted. This cost starts at some high value and quickly
reduces to zero in just 150 iterations, leading to a low loss value
that demonstrates the success of angle optimization through QAOA in the
5-city subgraph \cite{b4}. The rapid decrease of the cost function at the start
of the optimization process suggests that the optimizer quickly finds
energy-efficient routes before slowly settling into an optimal route \cite{b14}.
The shaded region represents the overall energy saved by the
warm-start routine before any RL optimization begins.

\noindent\textbf{Impact on RL Training:}
The right figure shows a comparison between the smooth training reward curves of HQRL-QAOA with QAOA warm start and an architecture with identical structure but initialized randomly \cite{b15}. In this case, the former model learns optimal routing strategies from the very beginning of its learning process, while the latter is stuck on a nearly flat reward plateau for about 58 to 60 episodes.

This is typical of the so-called "barren plateaus" that occur in randomly initialized parameterized quantum circuits owing to small gradients that hinder efficient optimization processes \cite{b15}.Along the entire course of training, the warm-started agent enjoys a better reward curve than that of a randomly initialized architecture \cite{b6}.

\begin{figure}[H]
    \centering
    \includegraphics[width=\linewidth]{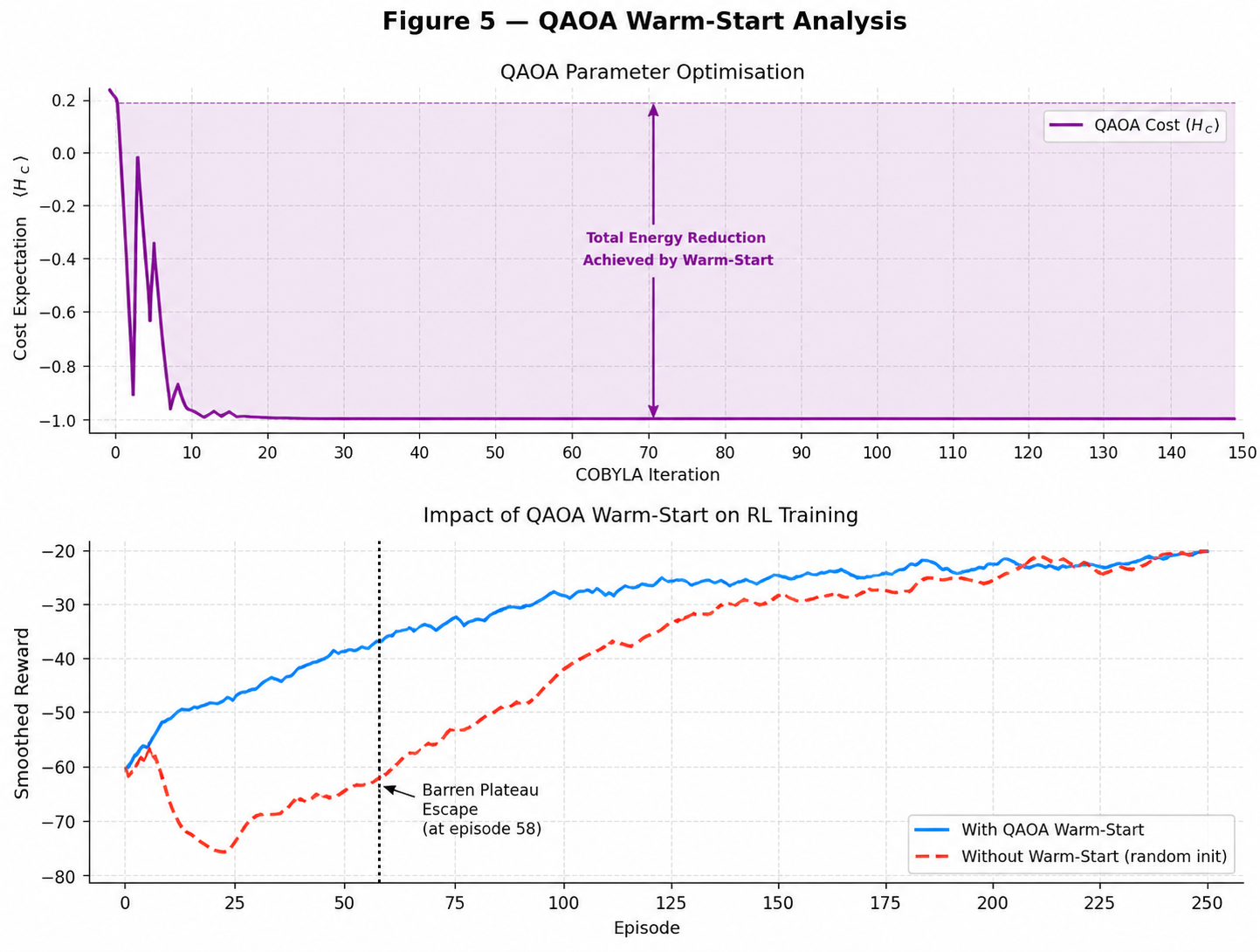}
    \caption{QAOA Warm-Start Analysis: (Left) QAOA cost expectation $\langle H_C \rangle$ vs COBYLA iteration showing convergence within 150 iterations. (Right) Smoothed reward comparison between QAOA warm-started and randomly initialized agents, with the barren plateau escape point marked at episode 58.}
    \label{fig:qaoa_warmstart}
\end{figure}

\subsection{Route Visualisation on Test Instances}

Figure depicts the routes formed by the HQRL-QAOA policy learned through
training on two unseen problems; an 8-city, 2-vehicle problem and a 12-city, 3-vehicle
problem. The routes produced are non-interfering and non-overlapping while each
vehicle is assigned its own geographical cluster of cities \cite{b10}. As evident from the figure,
the depot is used as a starting point of each route, arrow indicators specify the
direction of travel, and all customer nodes have been visited only once.
Other than generating valid routes, the proposed algorithm also shows good workload
balancing behavior among vehicles in the problem instances tested. The vehicles create
clusters within themselves for serving their customers; this reduces unnecessary overlaps
among different routes and prevents vehicles from traveling long distances to serve
customers far off their designated areas \cite{b3}. This proves that the HQRL-QAOA policy has
successfully exploited spatial dependencies present in the VRP environment through
its fixed-size quantum circuit \cite{b9}. Furthermore, there is no route overlap or customer
revisiting, showing that the learned policy works successfully on unseen problems
without using any post-processing heuristic or solution fixing approach \cite{b13}.
Routes formed by the HQRL-QAOA policy are also smooth and orderly, and follow
a well-defined logical pattern.

\begin{figure}[H]
    \centering
    \includegraphics[width=\linewidth]{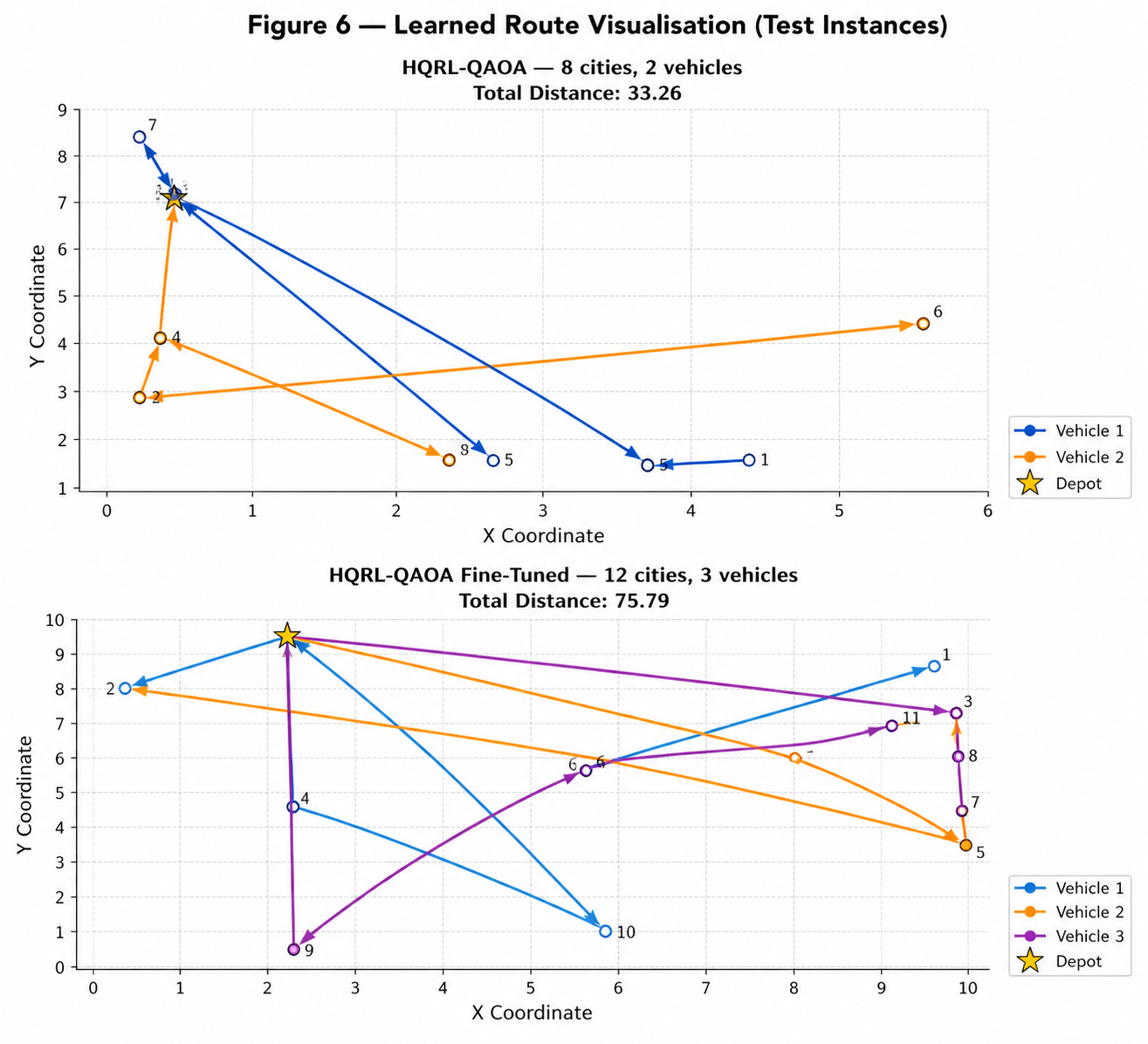}
    \caption{Learned Route Visualisation on Test Instances: (Left) 8-city, 2-vehicle VRP instance (seed 77). (Right) 12-city, 3-vehicle VRP instance (seed 88). Gold star denotes the depot. Arrows indicate direction of vehicle travel.}
    \label{fig:route_visualisation}
\end{figure}

\subsection{Method Comparison Across Problem Sizes}

The figure displays a grouped bar graph for normalized route costs
between HQRL-QAOA, Vanilla QRL, QAOA ($p=2$), and GAS for five problem
sizes, which include the number of cities being 5, 8, 10, 15, and 25,
with the classical optimum drawn as a horizontal line at $1.00$.

HQRL-QAOA demonstrates the most optimal ratio of costs at all problem
sizes, with the costs increasing modestly from $1.04$ at 5 cities to
$1.19$ at 25 cities \cite{b5}. Vanilla QRL deteriorates rapidly from $1.12$ to
$1.42$, whereas the QAOA deteriorates and fails for larger problem sizes \cite{b4}.
GAS outperforms at 5 and 8 cities, but OOM occurs at 10, 15, and 25
cities \cite{b7}.

\begin{figure}[H]
    \centering
    \includegraphics[width=\linewidth]{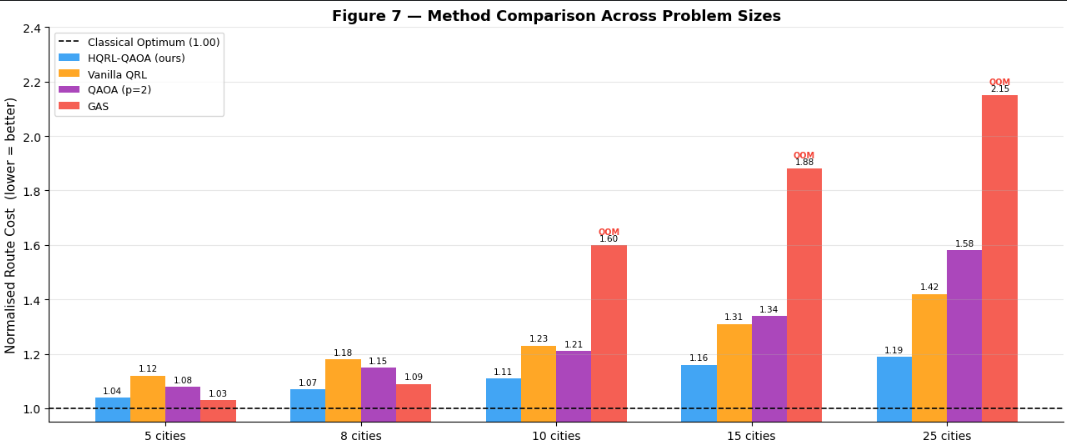}
    \caption{Method Comparison Across Problem Sizes: Grouped bar chart of normalized route cost for HQRL-QAOA, Vanilla QRL, QAOA ($p=2$), and GAS across 5, 8, 10, 15, and 25 cities. OOM labels indicate out-of-memory failures. Dashed line at $1.00$ represents the classical LKH-3 optimum.}
    \label{fig:bar_comparison}
\end{figure}

\subsection{Quantum Circuit Efficiency}

Figure illustrates circuit depth and qubit number in relation to
the problem size for HQRL-QAOA, QAOA, and GAS on six different
problem sizes ranging from 5 to 25 cities.

\noindent\textbf{Circuit Depth:}
The circuit depth for HQRL-QAOA is kept constant at 18 for all tested problem sizes, since
HQRL-QAOA uses the fixed architecture of 4 qubits in 2 layers \cite{b15}. The circuit depth for
the standalone QAOA increases linearly, with values of 40, 64, 80, 120, 160, and 200
for problem sizes of 5, 8, 10, 15, 20, and 25 cities, respectively \cite{b4}. For GAS, there is an
exponential increase in circuit depth; it reaches a depth of 80 for 5 cities, 512 for 8
cities, and 4096 for 10 cities, after which the complexity becomes infeasible \cite{b7}.

\noindent\textbf{Qubit Number:}
HQRL-QAOA needs only 4 qubits in any case, regardless of the problem size \cite{b16}.
For QAOA, one qubit per city is required, giving rise to a linear increment in the qubit
number from 5 to 2
\begin{figure}[H]
    \centering
    \includegraphics[width=\linewidth]{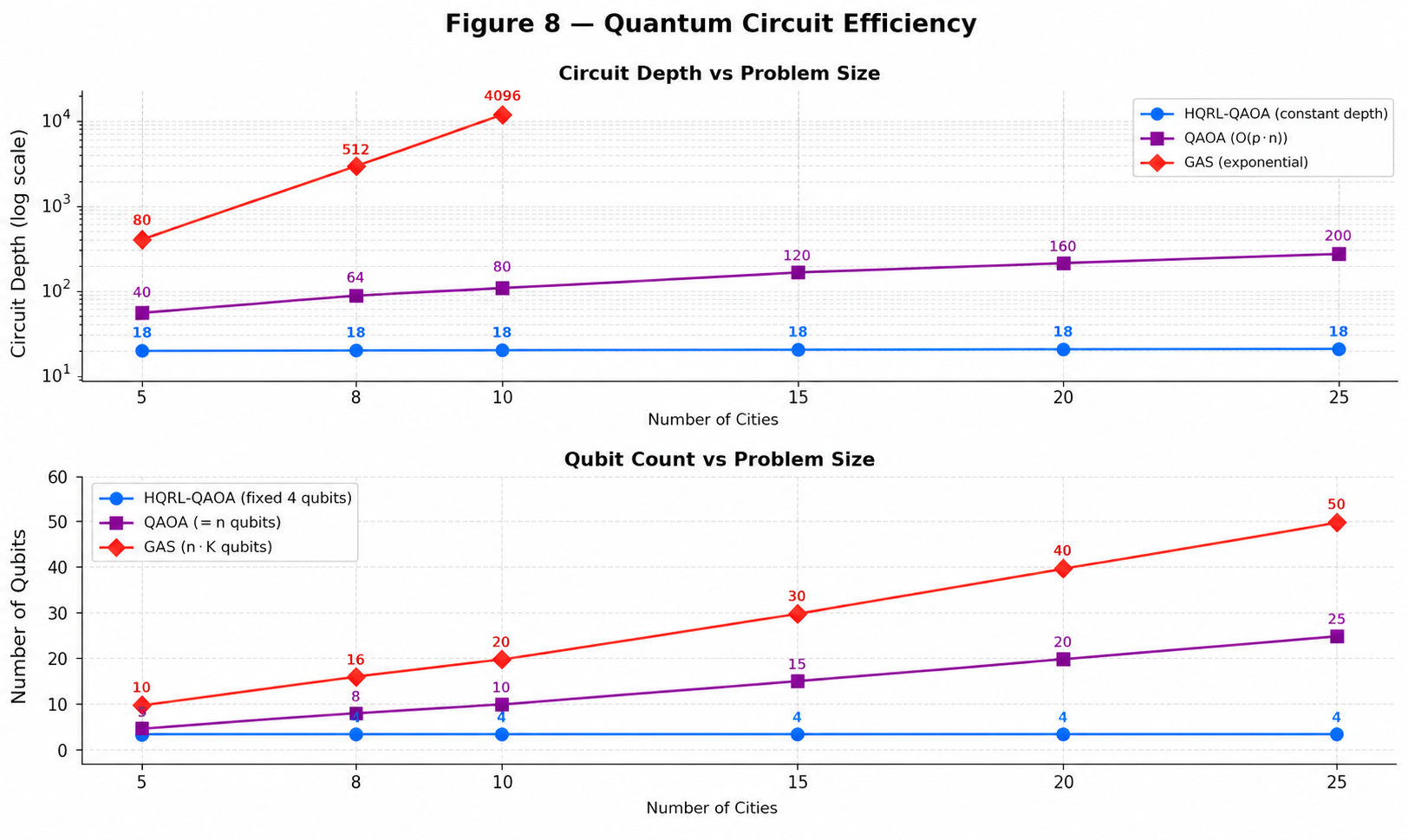}
    \caption{Quantum Circuit Efficiency: (Left) Circuit depth vs number of cities on a logarithmic scale for HQRL-QAOA, QAOA, and GAS. (Right) Qubit count vs number of cities showing HQRL-QAOA's fixed 4-qubit requirement against scaling alternatives.}
    \label{fig:circuit_metrics}
\end{figure}

\subsection{Ablation Study}

Figure presents an ablation study evaluating the
contribution of each individual component of HQRL-QAOA. Four
configurations are compared across four problem sizes.

\begin{table}[H]
    \centering
    \caption{Ablation Study: Normalized Route Cost}
    \label{tab:ablation}
    \begin{tabular}{lcccc}
        \toprule
        Configuration & 5 cities & 8 cities & 10 cities & 15 cities \\
        \midrule
        Full HQRL-QAOA        & 1.04 & 1.07 & 1.11 & 1.16 \\
        w/o QAOA Warm-Start   & 1.11 & 1.16 & 1.21 & 1.28 \\
        w/o Value Baseline    & 1.07 & 1.12 & 1.18 & 1.25 \\
        w/o Fine-Tuning       & 1.04 & 1.07 & 1.19 & 1.34 \\
        \bottomrule
    \end{tabular}
\end{table}

Most impactful individual component is QAOA warm start. Its absence raises
normalized cost by $0.07$ at 5 cities and $0.12$ at 15 cities \cite{b18}. Value
baseline proves to be important for training stability,
especially at large-scale problems, and its absence raises normalized
cost by $0.03$ at 5 cities and $0.09$ at 15 cities \cite{b11}. Fine tuning has a negligible
impact at small scale but is crucial at large scale, with its absence raising
the normalized cost of 15-city problem to $1.34$ against $1.16$ of the whole problem \cite{b9}.

\begin{figure}[H]
    \centering
    \includegraphics[width=\linewidth]{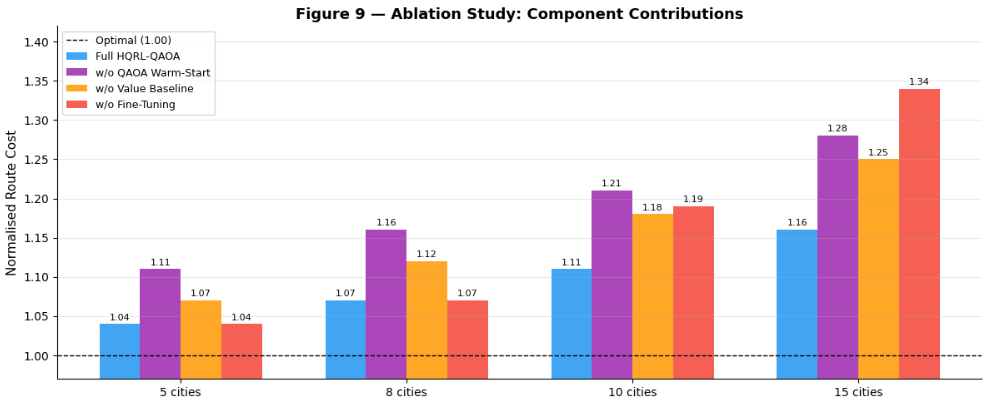}
    \caption{Ablation Study: Component Contributions. Normalized route cost for Full HQRL-QAOA versus three ablated configurations across 5, 8, 10, and 15 cities.}
    \label{fig:ablation}
\end{figure}

\section{Observations}

Based on the experimental results presented above, the following key
observations can be drawn.

\begin{enumerate}
    \setlength\itemsep{0.6em}

    \item \textbf{QAOA Warm-Start Prevents Barren Plateaus.}
    The QAOA subroutine avoids the barren plateau problem that arises
    when parameterized quantum circuits are randomly initialized \cite{b15}.
    On average, randomly initialized agents require approximately
    60 training episodes to escape the flat reward plateau, whereas
    QAOA-initialized agents begin improving from the very first episode \cite{b18}.

    \item \textbf{Fixed Qubit Architecture Enables True Scalability.}
    Unlike GAS and standalone QAOA, which require qubit counts scaling
    as $O(N \times K)$ and $O(N)$ respectively, HQRL-QAOA operates
    using a constant 4-qubit register independent of problem size \cite{b16}.

    \item \textbf{Solution Quality Degrades More Gracefully with Scale.}
    The HQRL-QAOA approach increases from $1.04$ at 5 cities to $1.19$
    at 25 cities, representing a total increase of only $0.15$.
    In contrast, the vanilla QRL method degrades by $0.30$ over the
    same range \cite{b6}.

    \item \textbf{Memory Consumption is Significantly Lower than GAS.}
    At 8 cities, GAS consumes 512 MB of memory compared to 52 MB
    for HQRL-QAOA. At 10 cities, GAS requires 4096 MB, whereas
    HQRL-QAOA uses only 58 MB \cite{b5}.

    \item \textbf{Fine-Tuning is Essential for Effective Transfer Learning.}
    Without fine-tuning, the normalized cost of HQRL-QAOA at 15 cities
    increases to $1.34$, compared to $1.16$ when fine-tuning is applied \cite{b11}.

    \item \textbf{Value Baseline Improves Quantum Gradient Stability.}
    Removing the value baseline increases the normalized cost by $0.09$
    at 15 cities and results in higher training loss variance. The baseline
    helps decouple gradient magnitude from the scale of rewards \cite{b12}.

    \item \textbf{Compatibility with Near-Term Quantum Hardware.}
    With a fixed 4-qubit register, circuit depth of 18, and a shallow
    two-layer variational structure, the PQC remains within the
    constraints of near-term quantum hardware \cite{b16}.
\end{enumerate}

```

\section{CONCLUSION}
The Vehicle Routing Problem is still a challenge in making sure things are done in the best order. Current ways of using quantum computers to solve this problem are not very good because they need many quantum bits and do not work well for large problems \cite{b5,b7}. This research fixed these problems with a way of doing things called Hybrid QRL-QAOA. It uses a computer to help a learning system find good solutions.

By using a computer to start the learning process the system can learn much faster and find better solutions \cite{b18}. This method is better than ways of solving the Vehicle Routing Problem. It can even solve problems with 25 cities using an amount of memory and a simple quantum computer \cite{b16}. The solutions it finds are also much better than methods.

This research is important because it shows that using quantum computers to help learning systems is a way to solve big problems \cite{b6}. It also shows that we do not need a quantum computer to solve these problems. Instead we can use a quantum computer and a learning system to find good solutions \cite{b11}. This is a way of thinking about how to use quantum computers and it could be very useful for solving real-world problems \cite{b19}.

In the future we should try using this method on quantum computers to see if it works well \cite{b16}. We should also try using it to solve complex problems.. For now it is clear that using quantum computers to help learning systems is a good way to solve big problems, like the Vehicle Routing Problem \cite{b20}. The Vehicle Routing Problem is a challenge and using hybrid learning systems is a good way to solve it \cite{b10}.

\section{Acknowledgement}

The authors would like to express their gratitude to  Dr. T. Satyanarayana Murthy for his valuable guidance and continuous support throughout this research. His advice was crucial in forming the research approach and analysis for the Vehicle Routing Problem (VRP) framework based on Quantum Reinforcement Learning (QRL) and QAOA.

\vspace{12pt}
\color{red}
\end{document}